\documentclass[conference]{IEEEtran}
\IEEEoverridecommandlockouts
\usepackage{amsmath,amssymb,amsfonts}
\usepackage{algorithmic}
\usepackage{graphicx}
\usepackage{textcomp}
\usepackage{xcolor}
\usepackage{xspace}
\usepackage{enumitem}
\usepackage{pifont}
\usepackage{bbding}
\usepackage{algorithm}
\usepackage{multirow}
\usepackage{caption}
\usepackage{authblk}
\usepackage{flushend}

\usepackage{setspace}
\renewcommand\baselinestretch{0.92}

\usepackage[square,sort,comma,numbers,compress]{natbib}

\begin{document}

\title{Dancing along Battery: Enabling Transformer with Run-time Reconfigurability on Mobile Devices}

\def\framework{RT$^3$\xspace}

\author[1]{Yuhong Song}
\author[2]{Weiwen Jiang}
\author[3]{Bingbing Li}
\author[1]{Panjie Qi}
\author[1, *]{Qingfeng Zhuge}
\author[1]{Edwin Hsing-Mean Sha}
\author[4]{\\Sakyasingha Dasgupta}
\author[2]{Yiyu Shi}
\author[3]{Caiwen Ding}

\affil[1]{Department of Computer Science \& Technology, East China Normal University, Shanghai, China}
\affil[2]{Department of Computer Science \& Engineering, University of Notre Dame, South bend, USA}
\affil[3]{Department of Computer Science \& Engineering, University of Connecticut, Mansfield, USA}
\affil[4]{Edgecortix Inc, Tokyo, Japan \vspace{-1em}}

\renewcommand\Authands{ and }

\maketitle

\begin{abstract}
A pruning-based AutoML framework for run-time reconfigurability, namely \framework, is proposed in this work. This enables Transformer-based large Natural Language Processing (NLP) models to be efficiently executed on resource-constrained mobile devices and reconfigured (i.e., switching models for dynamic hardware conditions) at run-time.
Such reconfigurability is the key to save energy for battery-powered mobile devices, which widely use dynamic voltage and frequency scaling (DVFS) technique for hardware reconfiguration to prolong battery life.
In this work, we creatively explore a hybrid block-structured pruning (BP) and pattern pruning (PP) for Transformer-based models and first attempt to combine hardware and software reconfiguration to maximally save energy for battery-powered mobile devices. Specifically, \framework integrates two-level optimizations: First, it utilizes an efficient BP as the first-step compression for resource-constrained mobile devices; then, \framework heuristically generates a shrunken search space based on the first level optimization and searches multiple pattern sets with diverse sparsity for PP via reinforcement learning to support lightweight software reconfiguration, which corresponds to available frequency levels of DVFS (i.e., hardware reconfiguration).
At run-time, \framework can switch the lightweight pattern sets within \textbf{45ms} to guarantee the required real-time constraint at different frequency levels.
Results further show that \framework can prolong battery life over $4\times$ improvement with less than 1\% accuracy loss for Transformer and 1.5\% score decrease for DistilBERT.
\end{abstract}


\section{Introduction}
The resurgence of deep learning boosted natural language processing (NLP) research through different deep neural networks (DNNs) such as simple Recurrent Neural Network (RNN) and Long Short-Term Memory (LSTM) unit \cite{hochreiter1997long}. 
Recently, the attention-based ``Transformer'' gains increasingly popularity due to its outstanding prediction accuracy. 
However, to approach human-level performance, these deeply layered ``Transformer'' aggressively adopt enormous model size, such as Turing-NLG with 17 billion parameters \cite{turing}.

In contrast, with the ongoing democratization of artificial intelligence (AI) \cite{garvey2018framework}, there is increasing needs to execute such giant models on mobile devices for low-latency and high-accuracy \cite{li2019edge,xu2018scaling}.
However, when Transformer-based models come to edge, the primary challenge is in accommodating giant models on resource-constrained mobile devices; along with the requirements to satisfy real-time performance.
Moreover, the battery-powered mobile devices naturally have a limited energy budget, which demands to prolong battery usage.

In order to tackle these challenges, we propose a novel two-level pruning-based AutoML framework, namely \framework, for efficiently executing the Transformer-based NLP tasks on battery-powered mobile devices with run-time reconfigurability to prolong battery usage.
\framework first applies a BP as the first-step compression to reduce model size. Then, it searches best multiple pattern sets (i.e., software reconfiguration) for PP using reinforcement learning (RL) and combines with DVFS (i.e., hardware reconfiguration) \cite{jiang2017optimal,jiang2016optimal}. It is worth noting that \framework is the first attempt on hybrid BP and PP for Transformer-based models and combining hardware and software reconfiguration.
Benefit from the regularity of BP, 
it is compatible with parallel computation used on mobile platforms.
Moreover, BP assists in building a shrunken search space of pattern sets heuristically, which avoids redundant searching and training cost for RL, so we call BP a ``hot'' search start.
Besides, the model after BP will be fixed as a backbone model for the second level optimization, that is to say, all pattern sets searched from RL will operate on the same model. 
As such, \framework can provide a lightweight pattern sets switch at run-time.
It should be noted that the run-time reconfigurability is not only applicable for DVFS, but can be applied for diverse scenarios, such as local language translation for on-line interactive events with a fluctuating network bandwidth.

The main contributions of this paper are listed as follows. 
\begin{itemize}[leftmargin=10pt]
\item As the first attempt, we propose to use a hybrid BP and PP for giant Transformer-based models, and retain accuracy, pruning rate and hardware efficiency simultaneously.
\item To prolong battery lifetime best and satisfy real-time constraint, we combine hardware and software reconfiguration, which has not been discussed in prior works.
\item We integrate a two-level optimization framework for run-time reconfigurability, namely \framework
, which is extensively evaluated. 
Experimental results shown that: \framework can prolong battery lifetime for over $4.9\times$ with less than $2\%$ accuracy loss, while guaranteeing the required timing constraints. 
\end{itemize}

\section{Related Work and Motivation}
\subsection{Related Work} 

For hardware-friendly weight pruning, work \cite{Shi_2020} divided RNN weight matrix into blocks and do row/column pruning in every block. \cite{li2020_efficient} applied such a hardware-friendly BP to Transformer-based models. However, this BP will result in a significant accuracy loss when pruning rate increases or block division is coarse-grained. In addition, \cite{ma2020pconv} presented PP method to make a better balance on accuracy and pruning rate. But it has not yet been implemented on Transformer-based models and the combination space of patterns may be too huge to find the best one when pattern size is large. In this work, we innovatively use a hybrid weight pruning combining BP and PP to trade-off accuracy, pruning rate and hardware efficiency.

On the other hand, there are works about run-time reconfigurability of DNN. \cite{yang2020nonuniform} generated switched convolutional neural networks (CNNs) based on an optimization problem.
\cite{sukhbaatar2019_adaptive} proposed a self-attention mechanism to learn adaptive attention span on Transformer-based models. 
However, existing works always focus on the selection of sub-models on software optimization, but ignore the hardware efficiency (e.g., inference latency and power consumption). Meanwhile, the granularity of compression method is coarse-grained (e.g., channel-wise compression or layer-wise compression) in sub-models generation process which will result significant accuracy degradation among sub-models. Besides, there is works using AutoML to identify a sub-model involving hardware feedback like \cite{wang2020hat,yang2020co-NOC,yang2020co-asic,jiang2019achieving,zhang2019neural,lu2019neural,jiang2019accuracy,jiang2018heterogeneous,jiang2020hardware,bian2020nass,jiang2020device,jiang2020standing}, but they focus on computer vision task or lack run-time hardware and software reconfiguration.

In general, \framework intelligently combines BP and PP to promote accuracy, pruning rate and hardware efficiency overall. Moreover, it implements both hardware and software reconfiguration to prolong battery usage best, which is not presented simultaneously in prior works. 


\subsection{Challenges and Motivation} 
\noindent{\textbf{Challenge 1: Cannot be accommodated on mobile devices.}}

In order to accommodate the large-size models to mobile devices with limited storage and computation resources,
model compression is essential; in particular the weight pruning.
The well-study non-structured pruning \cite{han2016deep_compression} and structured pruning \cite{wen2016learning} have their own defect.
The non-structured pruning that can prune any weights in the matrices leads to irregular sparsity, so the computation needs to rely on additional indices, which makes it
hard to be accelerated on current DNN accelerator.
On the other hand, the structured pruning can maintain a regular matrix with reduced dimensions but it suffers from notable accuracy loss due to coarse-grained pruning.

\vspace{3pt}
\noindent\textbf{Motivation 1: Adequate model pruning is highly demanded.} 
This work proposes a hybrid BP and PP for Transformer. BP is hardware-friendly, but it will suffer significant accuracy loss if pruning rate increases or block division is coarse-grained. Also, in this work, block size can't be set small for efficiency because blocks need to successively choose the best pattern from the pattern set. However, if PP is implemented solely with large block size, the pattern space is too huge to search. As a result,  
\framework employs BP (see Section \ref{sec:level1}) as the first-step pruning to keep hardware efficiency and assist PP to shrink the pattern space skillfully, and utilizes PP (see Section \ref{sec:level2}) to do further pruning and maintain the accuracy. 
Therefore, this hybrid pruning can hold accuracy, pruning rate and hardware efficiency simultaneously.

\vspace{3pt}
\noindent\textbf{Challenge 2: Battery-powered device has limited energy.}

\begin{table}[t]
\setlength{\abovecaptionskip}{0.1cm}
\setlength{\belowcaptionskip}{-0.6cm} 
  \centering
  \begin{footnotesize}
  \tabcolsep 6pt
  \renewcommand\arraystretch{1.1}
    \begin{tabular}{|c|cccccc|}
    \hline
     Notation & $l_1$ & $l_2$ & $l_3$ & $l_4$ & $l_5$ & $l_6$\\
    \hline
     freq (MHz) & 400 & 600 & 800 & 1000 & 1200 & 1400 \\
     vol (mV) & 916.25 & 917.5 & 992.5 & 1066.25 & 1141.25 & 1240 \\
    \hline
    \end{tabular}%
  \end{footnotesize}
  \captionsetup{font={footnotesize}}
    \caption{Illustration of Voltage/Frequency levels supported by ARM Cortex A7 core in Odroid-XU3 mobile platform}
  \label{tab:volfre}%
\end{table}%

Most mobile devices are battery-powered, such as mobile phone, unmanned aerial vehicle (UAV), and robotic \cite{shi2016promise}.
It is important to prolong battery usage of devices with limited energy;
meanwhile, the tasks need to comply with real-time requirements.
DVFS \cite{horowitz1994low} is a promising and widely used technique to prolong the battery life in each charging cycle.
Table \ref{tab:volfre} demonstrates the available Vlotage/Frequency (V/F) levels for the Cortex A7 core in Odroid-XU3. When remained energy becomes low, the running frequency is scaled down to enter an energy-saving mode (e.g., iPhone can turn into energy-saving mode when the battery is lower than 20\%).

We do experiments to demonstrate that DVFS can prolong the battery usage, results show in Table \ref{tab:mot1}.
In this table, all approaches have the same energy budget.
Both approaches E1 and E2 apply the same Transformer model $M1$, approach E2 has three DVFS modes: F-Mode for fast execution; N-Mode for normal-speed execution; and E-Mode for energy-saving execution, while E1 has no DVFS.
Results show that E2 can achieve 17.30\% improvement than the number of runs for model $M1$.
However, when running frequency is scaled down, the N-Mode and E-Mode of E2 cannot guarantee to satisfy the real-time constraint of $115ms$.

To overcome this problem, the software reconfiguration is explored to adapt the DVFS, which is demonstrated in approach E3 in Table \ref{tab:mot1}.
With both hardware reconfiguration (i.e., DVFS) and software reconfiguration (i.e., multiple pattern sets with diverse sparsity), it can satisfy the timing constraint (i.e., $115ms$), and significantly improve the number of runs of models (i.e., E3 is 1.78$\times$ times than E1).
As a penalty, it may make a degree of sacrifice on accuracy among sub-models.

\vspace{3pt}
\noindent\textbf{Motivation 2: Coupling DVFS with software reconfiguration to fully use the limited battery energy.} Now, it is clear that without software reconfiguration, the real-time requirements cannot be satisfied at low V/F level.
Therefore, in the second level of \framework (see Section \ref{sec:level2}), we first heuristically build a shrunken search space based on BP, then we to select best pattern sets using RL and switch according to available V/F levels of DVFS to guarantee the real-time performance and prolong battery usage. Besides, we use both hardware and software metrics to guide the RL search.

\begin{table}[t]
\setlength{\abovecaptionskip}{0.1cm}
\setlength{\belowcaptionskip}{-0.6cm} 
  \centering
  \tabcolsep 3pt
  \renewcommand\arraystretch{1.1}
    \begin{footnotesize}
    \begin{tabular}{|cccccccc|}
    \hline
    \multirow{2}{*}{App.} & \multicolumn{2}{l}{\textbf{Software}} & \multicolumn{5}{l|}{\textbf{Hardware}} \\
          & Model & Accuracy & DVFS  & Lat. (ms) & Sat.  & \# runs & Imp \\
    \hline
    E1    & M1    & 96.81\% & F-Mode & 114.59 & \checkmark & $1.53\times10^6$ & - \\
    \hline
    \multirow{3}[0]{*}{E2} & \multirow{3}[0]{*}{M1} & \multirow{3}[0]{*}{96.81\%} & F-Mode & 114.59 & \checkmark     & \multirow{3}[0]{*}{$1.85\times10^6$} & \multirow{3}[0]{*}{17.30\%} \\
          &       &       & N-Mode & 160.43 & $\times$     &       &  \\
          &       &       & E-Mode & 200.54 & $\times$     &       &  \\
    \hline
    \multirow{3}[1]{*}{E3} & M1    & 96.81\% & F-Mode & 114.59 & \checkmark     & \multirow{3}[1]{*}{$2.72\times10^6$} & \multirow{3}[1]{*}{$1.78\times$} \\
          & M2    & 96.26\% & N-Mode & 100.54 & \checkmark     &       &  \\
          & M3    & 93.57\% & E-Mode & 90.62 & \checkmark     &       &  \\
    \hline
    \end{tabular}%
    \end{footnotesize}
    \captionsetup{font={footnotesize}}
    \caption{Comparison of three different approaches and Transformer models (Software, SW) running on hardware (HW) with the timing constraint of $115ms$: E1 without any run-time reconfiguration, E2 with only HW reconfiguration, E3 with both SW and HW reconfiguration.}
  \label{tab:mot1}%
\end{table}%


\subsection{Problem Definition}
This paper aims to address all the above challenges in a holistic framework.
We define the problem as follows: Given a pre-trained Transformer model $M$, a set of available run-time V/F levels $L$, an energy budget $E$, a timing constraint $T$, the problem is to determine: (i) $C:$ model based on BP; (ii) $P:$ multiple pattern sets produced by RL to enable software reconfiguration.
such that, each execution of model $M$ under model pruning (i.e., BP and PP) at a V/F level controlled by remained energy can satisfy timing constraint $T$, meanwhile, the number of runs of model $M$ can be maximized. 
Kindly note that the number of runs indicates the number of inference within the energy budget, it reflects the efficiency of battery usage. In the following of this paper, we will use the metrics ``\textit{number of runs}'' for hardware efficiency evaluation.

\section{\framework: Run-Time Re-configurable Transformer with Real-Time Performance}
\subsection{\framework~framework overview}

\begin{figure}[t]
\setlength{\abovecaptionskip}{0.1cm}
\setlength{\belowcaptionskip}{-0.6cm} 
\centering
\includegraphics[width=1.0\linewidth]{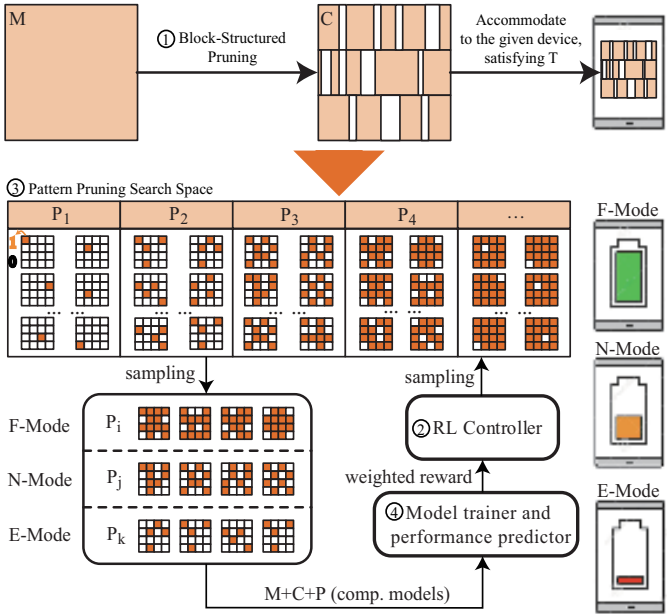}
\captionsetup{font={footnotesize}}
\caption{Overview of the proposed \framework~framework with two-level optimizations: (1) Block-structured pruning as the first-step pruning; (2) reinforcement learning to determine multiple pattern sets to support run-time reconfiguration in terms of DVFS. Kindly note that we use small block size of $4\times 4$ in this figure just for the ease of illustration, our experiments use $100\times 100$ to mitigate computation overhead.
}
\label{fig:overview}
\end{figure}

Figure \ref{fig:overview} illustrates the overview of the proposed \framework~framework.
\framework~is composed of two levels of optimization, which are further composed of four components: (in Level 1) \ding{172} BP is conducted on the original Transformer model (M) to reduce its size to a pruned model (C); (in Level 2)
\ding{173} the RNN-based RL controller guides the search process;
\ding{174} PP search space with multiple pattern sets which is heuristically built upon model (C); 
\ding{175} a model trainer to train the shared backbone model (C) using multiple sampled pattern sets and generate the reward and feedback to RL controller.
We will introduce each component specifically as follows.

\subsection{Level 1: First-step model pruning}\label{sec:level1}
In response to Challenge 1, we propose the hardware-friendly BP as the first-step pruning to keep matrix regularity and guarantee a degree of pruning rate and accuracy.

\vspace{3pt}
\noindent\textit{\ding{172} Block-structured pruning (BP)}
\vspace{2pt}

For irregular pruning~\cite{han2015learning}, the irregular sparse matrix is stored using Coordinate Format (COO), the nonzeros and related coordinates in memory need to be stored. Thus, three vectors are needed: row, col, data, where data[i] is the value at (row[i], col[i]) position. 
Our BP is conducted by excluding entire rows/columns within blocks, which just needs to store the row/column index and value of nonzeros. 
It significantly reduce the number of indices on memory storage. 

The weight pruning problem can be formulated using the reweighted group Lasso to orchestrate the BP. 
As shown in the matrix C in Figure~\ref{fig:overview}, 
we divide the weight matrix M into 3 row-wise \emph{blocks} and apply the \emph{column pruning} on each block.
Kindly note that we use column pruning here as an example, it can be generalized to apply row pruning or both row and column pruning.
For each block, we compute the $l_2$ norm of each row/column. We prune the rows/columns according to our pre-set threshold or percentile, which is decided by lots of experiments. The pseudocode is shown in Algorithm~\ref{alg:block_pruning}.
Consider an $N$-layer Transformer, we denote the weights and biases of the $n$-th layer as $\mathbf{W}_n$ and $\mathbf{b}_n$. The loss function is $f \big( \{{\mathbf{W}}_{n}\}_{n=1}^N, \{{\mathbf{b}}_{n} \}_{n=1}^N \big)$, which will be minimized during training. For BP, our objective is to reduce the number of rows/columns in the blocks while maintaining the accuracy.

\renewcommand{\algorithmicrequire}{\textbf{Input:}}
\setlength{\intextsep}{3pt}
\begin{algorithm}[H]
\footnotesize
\captionsetup{font={small}}
  \caption{Block-structured pruning }
  \label{alg:block_pruning}
\begin{algorithmic}[1]
   \REQUIRE{weight matrix $\mathbf{W}$, row division $k$ (or column division $k'$), pre-set $l_2$ norm threshold $t_b$}
  \STATE Divide $\mathbf{W}$ into $k$/$k'$ row/column-wise blocks in row/column dimension
  \FOR{every row/column-wise blocks $BW_i$}
    \FOR{calculate $l_2\_normj$ of each row/column $j$}
    \IF{$l_2\_normj$ lower than $t_b$}
        \STATE{remove the whole row/column $j$}
    \ENDIF
    \ENDFOR
  \ENDFOR
\end{algorithmic}
\end{algorithm}


Compared with irregular pruning, BP can achieve better hardware efficiency. 
However, in order to maintain accuracy, the pruning rate will be limited when the block size increases. Therefore, we propose PP in the second level to prune model further. Except to keep regular sparse model and guarantee the accuracy, BP can also narrow the search space of RL to cut down searching and training cost. Besides, model after BP as a fixed backbone model provides a lightweight pattern sets switch to support run-time reconfiguration.
 

\subsection{Level 2: Enabling Run-Time Reconfiguration}\label{sec:level2}

Level 2 in \framework~is the key to enable run-time reconfigurability.
In response to challenges 2, we take the output model (C) of Level 1 as a fixed backbone model, and we employ PP (i.e., pruning using pattern sets) to provide software flexibility for different V/F levels.
As such, we can lock the fixed backbone model and switch pattern sets only for reconfiguration.
We will introduce the details in components \ding{173}-\ding{175} to demonstrate how to support reconfiguration of Transformer.

\vspace{3pt}
\noindent\textit{\ding{173} Reinforcement learning controller}
\vspace{2pt}

The RL controller is implemented based on an RNN, similar to \cite{zoph2016neural}.
The RL controller predict $N$ pattern sets based on a softmax classifier from \ding{174} PP search space for $N$ V/F levels.
Then, the controller further predict $K$ patterns from each selected pattern set.
The parameters $\theta_c$ in RNN will be updated using the metrics from component \ding{175}.
A policy gradient method will be employed to update parameters $\theta_c$ to predict better architectures over a series of episodes.

In each episode, the predicted patterns can be regarded as actions.
Based on these actions, the pattern sets for different V/F levels can be determined.
At the end of an episode, a reward is sent back to the controller. The reward is computed according to the following procedures: (1)  (from \ding{175}) calculate the latency $lat_i$ and number of runs $runs_i$ for each V/F level $l_i$;
(2) verify whether timing constraint $T$ can be satisfied; if $\exists lat_i > T$, we directly calculate the reward without fine-tuning the model. otherwise, 
(3) fine-tune the backbone model to obtain accuracy $acc_i$ for the $i^{th}$ pattern set on a hold-out dataset.
After these steps, we utilize both the number of runs $runs_i$ (i.e., hardware metric) and accuracy $acc_i$ (i.e., software metric) for the $i^{th}$ pattern set to guide the search.

With the obtained information, we can formulate the reward function.
Before introducing the detailed function, we first define five notations: (1) $A_w$: the weighted accuracy associated to all pattern sets, which is determined as $A_w=\sum\nolimits_{\forall i}\{\alpha_i\times acc_i\}$; (2) $A_o$: the accuracy of model (C) output from Level 1; (3) $A_m$: a pre-set lowest accuracy; (4) $cond$: a binary condition, $cond=True$ if for $\forall i<j$ we have $acc_i>acc_j$; otherwise we have $cond=False$, indicating that the model $M_i$ for low V/F level $l_i$ has higher accuracy over model $M_j$.
In this case, 
we give a penalty $pen$ to the reward. (5) The reward for the number of runs $R_{runs}$ is normalized to the range of $[0,1]$.
Based on these definitions, we formulate reward $R$ as follows:

\begin{equation}
\setlength{\abovedisplayskip}{-5pt}
\setlength{\belowdisplayskip}{-10pt}
\begin{small}
R=\left\{
\renewcommand\arraystretch{1.5}
\begin{array}{ll}
-1 + R_{runs} & {\exists lat_i > T}\\
\frac{A_w-A_{m}}{A_{o}-A_{m}}+ R_{runs}  & {\forall lat_i \le T\ \& \ cond=True}\\
\frac{A_w-A_{m}}{A_{o}-A_{m}}-pen+ R_{runs} & {otherwise}\\
\end{array} \right.
\end{small}
\end{equation}
There are three cases in calculating reward $R$: (1) if $\exists lat_i > T$ indicating timing constraint cannot be met at least one pattern set, we directly set $R$ to -1+$R_{runs}$ and will not fine-tune the model; (2) if $\forall lat_i \le T$ and $cond=True$, we use the normalized accuracy and number of runs as reward; finally, (3) if $cond=False$, we give a penalty on the reward.

\vspace{3pt}
\noindent\textit{\ding{174} Generating pattern pruning search space}
\vspace{2pt}

PP has been widely used in CNNs \cite{niu2020patdnn}; however, it is non-trivial to study how to apply PP to Transformer-based models.
The main challenge comes from the much larger dimension of weights, which even reaches $28785\times 800$ in our experiments. And the Transformer don't have a \textit{``kernel"} component to decide pattern size naturally.
A small pattern will lead to computation overhead, while a large pattern suffers from the low accuracy.
For a better efficiency-accuracy trade-off, the pattern size ($p_{size}\times p_{size}$) is set to $100\times 100$ in this work, which leads to $10^4$ different pattern sets with diverse sparsity.
In addition, the number of patterns with the same sparsity can be as large as $C(100,50)=8.6\times10^{286}$.
Obviously, it is impossible to search from all the possible patterns.

To solve the above problem, we first use the input constraint to select pattern sets. Given $N$ V/F modes and the timing constraint $T$, we can predict the $N$ sparsity ratios nearest to $T$.
Then, we gradually tight the constraints to involve $\theta\times N$ sparsity ratios in total.
Second, we skillfully take use of the obtained model (C) from Level 1 to select $m$ representative patterns for each pattern set.
The selection process is conducted as follow.
Considering that $C$ can be divided into $n$ square blocks, we sample $\frac{n}{2}$ blocks and conduct point-wise addition on these blocks, which can obtain the importance weight for each position in a $p_{size}\times p_{size}$ square.
Then, according to the sparsity ratio, we set 0 in the pattern for all less important weights, and we set all other positions to be 1.
In this way, one pattern is constructed.
The construction procedure will repeat $m$ times to create a pattern set.

\vspace{3pt}
\noindent\textit{\ding{175} Model trainer and performance predictor}
\vspace{2pt}

\begin{figure}[t]
\setlength{\abovecaptionskip}{0.1cm}
\setlength{\belowcaptionskip}{-0.6cm} 
\flushleft
\includegraphics[width=1.0\linewidth]{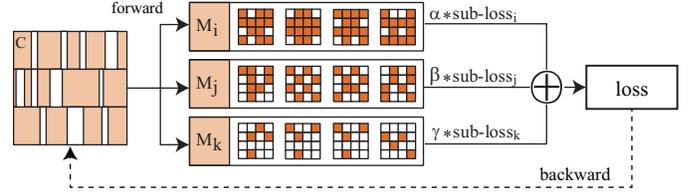}
\captionsetup{font={footnotesize}}
\caption{Off-line training of shared backbone model: (1) forward propagation: go through each pattern set to obtain the sub-loss, and then accumulate the weighted sub-loss to obtain the loss; (2) backward propagation: update the weight of model $C$ using the accumulated loss.}
\label{fig:joint_training}
\end{figure}

Figure \ref{fig:joint_training} demonstrates the jointly training procedure to obtain the shared backbone model based on model $C$.
In forward propagation, $C$ will go through each pattern set to get the sub-loss.
Like \cite{Shi_2020}, $C$ is divided into blocks with the same size of pattern, and we choose the pattern with the largest l2-norm for each block.
We accumulate all sub-loss to obtain a weighted loss, which will be used to update the model in backward propagation.
The forward and backward propagation will be performed in $\xi$ epochs to train the shared backbone model, and the accuracy $acc_i$ for the $i^{th}$ pattern set can be obtained by executing one more time of forward propagation.

The performance predictor will predict the inference latency. 
It is obvious that applying different pattern sets with diverse sparsity for the shared model will lead to different latency.
We apply the same compiler technique in \cite{niu2020patdnn} to optimize the PP, which can also predict the execution clock cycles. Thus, the latency $lat_i$ using $i^{th}$ pattern set on the backbone model for a V/F level can be obtained. Next, the number of runs $R_{runs}$ can be calculated based on the latency and battery energy $E$.

\begin{table*}[htbp]
\setlength{\abovecaptionskip}{0.1cm}
\setlength{\belowcaptionskip}{-0.3cm} 
  \centering
  \begin{footnotesize}
  \tabcolsep 5.2pt
  \renewcommand\arraystretch{1.1}
    \begin{tabular}{|cc|ccc|ccc|ccc|ccc|ccc|}
    \hline
    \multicolumn{2}{|c|}{Dataset/Task} & \multicolumn{3}{c|}{WikiText-2 (T: 94ms)} & \multicolumn{3}{c|}{WikiText-2 (T: 104ms)} & \multicolumn{3}{c|}{RTE (T: 200ms)} & \multicolumn{3}{c|}{STS-B (T: 330ms) } \\
    \hline
    \multicolumn{2}{|c|}{\multirow{2}[1]{*}{Models}} & \multicolumn{3}{c|}{Transformer} & \multicolumn{3}{c|}{Transformer} & \multicolumn{3}{c|}{DistilBERT} & \multicolumn{3}{c|}{DistilBERT}\\
    \multicolumn{2}{|c|}{} & M1 & M2 & M3 & M1 & M2 & M3 & M1 & M2 & M3 & M1 & M2 & M3 \\
    \hline
    \multicolumn{2}{|c|}{Sparsity} & 70.80\% & 80.61\% & 87.32\% & 73.95\% & 77.44\% & 83.83\% & 51.78\% & 67.52\% & 85.96\% & 42.98\% & 47.73\% & 54.83\% \\
    \multicolumn{2}{|c|}{Latency (ms)} & 93.55  & 86.78  & 70.72  & 83.40  & 101.06  & 90.31  & 199.94  & 188.58  & 101.92  & 236.44  & 303.43  & 327.73  \\
    \hline
    \multirow{2}[1]{*}{UB} & Accuracy & 97.27\% & 96.29\% & 93.03\% & 97.36\% & 97.31\% & 96.58\% & 58.12\% & 56.32\% & 53.43\% & \multicolumn{1}{r}{82.68\%} & \multicolumn{1}{r}{79.45\%} & \multicolumn{1}{r|}{70.84\%} \\
      & Interrupt & \multicolumn{3}{c|}{51.82 seconds} & \multicolumn{3}{c|}{51.82 seconds} & \multicolumn{3}{c|}{66.93 seconds} & \multicolumn{3}{c|}{66.94 seconds}\\
    \hline
    \multirow{2}[1]{*}{\textbf{\framework}} & \textbf{Accuracy} & \textbf{95.40\%} & \textbf{95.37\%} & \textbf{90.04\%} & \textbf{97.13\%} & \textbf{97.00\%} & \textbf{95.36\%} & \textbf{55.60\%} & \textbf{54.51\%} & \textbf{52.71\%} & \multicolumn{1}{r}{\textbf{82.09\%}} & \multicolumn{1}{r}{\textbf{78.30\%}} & \multicolumn{1}{r|}{\textbf{69.51\%}}\\
      & \textbf{Interrupt} & \multicolumn{3}{c|}{\textbf{8.75 milliseconds}} & \multicolumn{3}{c|}{\textbf{8.75 milliseconds}} & \multicolumn{3}{c|}{\textbf{44.90 milliseconds}} & \multicolumn{3}{c|}{\textbf{45.00 milliseconds}} \\
    \multicolumn{2}{|c|}{\textbf{Accuracy gap}} & \textbf{1.87\%} & \textbf{0.92\%} & \textbf{2.99\%} & \textbf{0.23\%} & \textbf{0.31\%} & \textbf{1.22\%} & \textbf{2.53\%} & \textbf{1.81\%} & \textbf{0.72\%} & \textbf{0.60\%} & \textbf{1.15\%} & \textbf{1.33\%}\\
    \hline
    \end{tabular}%
    \end{footnotesize}
    \captionsetup{font={footnotesize}}
    \caption{AutoML results for Transformer and DistilBERT on WikiText-2 dataset and RTE and STS-B tasks in GLUE.}
  \label{tab:exp0}%
\end{table*}%

\begin{figure*}[t]
\setlength{\abovecaptionskip}{0.1cm}
\setlength{\belowcaptionskip}{-0.6cm} 
\centering
\includegraphics[width=1\linewidth]{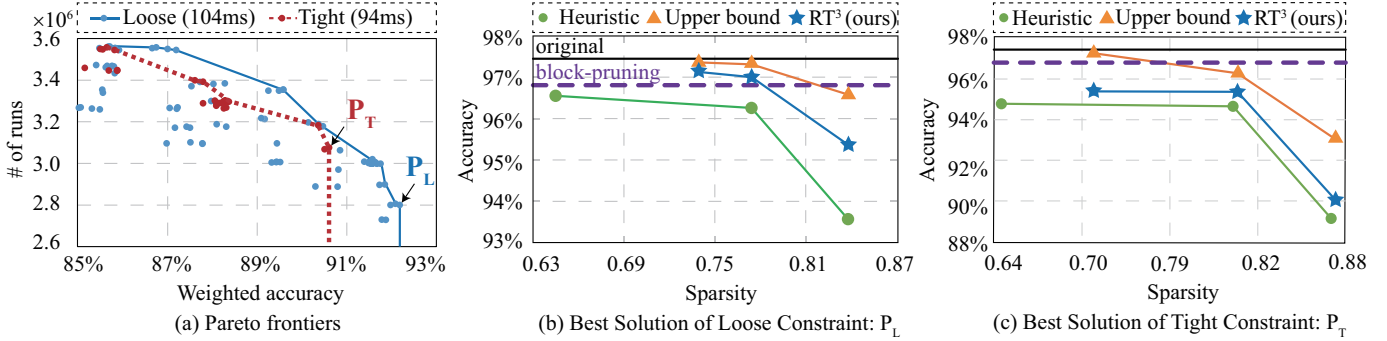}
\captionsetup{font={footnotesize}}
\caption{Search space exploration: (a) results of Transformer under the loose (104ms) and tight (94ms) timing constraints; (b) the best solution ($P_L$) obtained by \framework using loose timing constraint; (c) the best solution ($P_T$) for tight timing constraint.}
\label{fig:exp1}
\end{figure*}

\section{Experiment and Comparison}
\subsection{Experimental setup}

\textbf{Baseline Models and Datasets.} We test our method on Transformer using WikiText-2 dataset\cite{merity2016pointer} and DistilBERT models on GLUE benchmark \cite{wang2019glue}, a comprehensive collection of nine natural language understanding tasks. We conduct the experiments using Pytorch Transformer model and HuggingFace Transformer toolkit for the state-of-the-art NLP \cite{wolf2019huggingface}.
For the transformer, there are two encoder and one decoder layers. For the DistilBERT, there are 6 encoder layers with hidden size $H$ = 768, number of heads $A$ = 12.


\textbf{Evaluation Metrics.} On the Transformer model, we use the accuracy of next word prediction. On DistilBERT in GLUE benchmarks, we report the metrics following the conventions in \cite{wang2019glue}, i.e., accuracy scores for SST-2, QNLI, RTE, and WNLI; Matthews Correlation Coefficient (MCC) for CoLA; F1 scores for QQP and MRPC; and Spearman correlations for STS-B. To demonstrate hardware efficiency, we use the number of runs to measure the usage of battery.

\textbf{Evaluation Platforms.} We conduct the AutoML optimization with the training of Transformer-based models on an 8$\times$ NVIDIA Quadro RTX 6000 GPU server (24 GB GPU memory). Experiments are performed on Python 3.6.10, GCC 7.3.0, PyTorch 1.4.0, and CUDA 10.1. 
For the target mobile device, we use the mobile platform Odroid-XU3 \cite{odroidxu3}.


\subsection{\framework: Search for Best Model}

This subsection evaluates the efficacy of \framework using Transformer on WikiText-2 dataset and applying DistilBERT on RTE and STS-B tasks.
We select 3 V/F levels (i.e., $\{l_3,l_4,l_6\}$ in Table \ref{tab:volfre}) for evaluation, and \framework searches for three sub-models (i.e., $\{M1, M2, M3\}$) correspondingly.

\textbf{Compare with Accuracy Upper Bound.}
The RL algorithm will identify three pattern sets, corresponding to $\{M1,M2,M3\}$.
If we train these models individually, the obtained accuracy will provide an upper bound (denoted as ``UB'') for comparison. An individual training means that three models and three pattern sets will be switched at run-time.
Table \ref{tab:exp0} reports the results.
First, all the models in \ref{tab:exp0} under pre-set timing constraint, i.e., 94ms and 104ms for WikiText-2, 200ms for RTE and 330ms for STS-B, which indicates that all the models satisfy real-time inference.
Second, The experimental results clearly show that the accuracy of \framework with jointly training provides acceptable accuracy loss than UB.
For example, in WikiText-2 ($T: 104ms$), the largest and average accuracy gap are 1.22\% and {0.59\%}, while for STS-B, these figures are 1.33\% and  {1.02\%}.
Third, \framework enables lightweight pattern sets switch compared with UB, which is intolerant because of more than 1 minute switch time. For DistilBERT, \framework achieves over $1000\times$ speedup at switch.

\textbf{Search space exploration.}
We collect the explored solutions from RL to form the search space exploration results in Figure \ref{fig:exp1} (a).
In this figure, the x-axis and y-axis stand for the weighted accuracy and the number of runs. 
As the same with the previous WikiText-2 experiments, we have two timing constraints: 94ms for the tight deadline, and 104ms for loose deadline.
This figure shows the Pareto frontier for these two timing constraints. 
It is clear, when we set a loose constraint, the Pareto frontier covers the one with tight constraint. This is reasonable since the tight constraint will require higher sparsity, leading to more accuracy loss.
In these Pareto frontiers, we select the ones ($P_T$ and $P_L$) with the highest accuracy and fine-tune the models to obtain the final results, as shown in Figure \ref{fig:exp1}(b)-(c), respectively.

\begin{table*}[htbp]
\setlength{\abovecaptionskip}{0.1cm}
\setlength{\belowcaptionskip}{-0.5cm} 
  \centering
  \renewcommand\arraystretch{0.75}
  \begin{footnotesize}
    \tabcolsep 15pt
    \renewcommand\arraystretch{1}
    \begin{tabular}{|c|c|c|c|c|c|c|c|}
    \hline
    Dataset/Task & Methods & No-Opt & rBP only & rBP+rPP & rBP+PP & BP only & \textbf{\framework } \\
    \hline
    \multirow{5}[6]{*}{WikiText-2} & Avg. Spar. & 0.00\% & 64.26\% & 77.75\% & 79.04\% & 64.26\% & \textbf{75.24\%} \\
\cline{2-8}      & \# runs($10^6$) & 0.55 & 1.52 & 3.58 & 3.19 & 1.53 & \textbf{2.71} \\
      & Impr. & - & 2.80$\times$  & 6.55$\times$  & 5.84$\times$  & 2.80$\times$  & \textbf{4.96$\times$ } \\
\cline{2-8}      & Avg. Acc & 97.45\% & 95.42\% & 86.38\% & 92.57\% & 96.81\% & \textbf{95.50\%} \\
      & Acc. loss & - & 2.03\% & 11.07\% & 4.88\% & 0.64\% & \textbf{0.95\%} \\
    \hline
    \hline
    \multirow{5}[6]{*}{RTE} & Avg. Spar. & 0.00\% & 49.31\% & 68.46\% & 68.35\% & 49.31\% & \textbf{68.42\%} \\
\cline{2-8}      & \# runs($10^6$) & 0.42 & 0.84 & 1.78 & 1.76 & 0.84 & \textbf{1.77} \\
      & Impr. & - & 1.97$\times$  & 4.19$\times$  & 4.16$\times$  & 1.97$\times$  & \textbf{4.17$\times$ } \\
\cline{2-8}      & Avg. Acc & 59.20\% & 58.48\% & 52.47\% & 53.79\% & 59.20\% & \textbf{54.27\%} \\
      & Acc. loss & - & 0.72\% & 7.09\% & 6.61\% & 0.00\% & \textbf{4.93\%} \\
    \hline
    \hline
    \multirow{5}[6]{*}{STS-B} & Avg. Spar. & 0.00\% & 39.47\% & 47.89\% & 47.87\% & 40.00\% & \textbf{48.51\%} \\
\cline{2-8}      & \# runs($10^6$) & 0.42  & 0.70  & 0.96  & 0.96  & 0.71  & \textbf{0.97 } \\
      & Impr. & - & 1.65$\times$  & 2.27$\times$  & 2.27$\times$  & 1.67$\times$  & \textbf{2.30$\times$ } \\
\cline{2-8}      & Avg. Acc & 86.50\% & 60.51\% & 50.17\% & 51.83\% & 83.70\% & \textbf{77.66\%} \\
      & Acc. loss & - & 25.99\% & 36.33\% & 34.67\% & 2.80\% & \textbf{8.84\%} \\
    \hline
    \end{tabular}%
    \end{footnotesize}
    \captionsetup{font={footnotesize}}
    \caption{Evaluation of the proposed block-structured pruning and AutoML based pattern pruning on WikiText-2 dataset and RTE and STS-B tasks.}
  \label{tab:ablation study}%
\end{table*}%

In Figure \ref{fig:exp1}(b)-(c), we further include a heuristic as a baseline approach for comparison. It selects the pattern sets with the sparsity that just satisfy the timing constraint for each V/F level, and jointly train for the backbone model.
In addition, we show the accuracy of the original model without compression and the backbone model obtained by BP.
From Figure \ref{fig:exp1}(b), we have an interesting observation that both UB and \framework can find solutions with accuracy higher than the BP.
This is possible because that \framework can conduct sparsity regularization induced by BP, like that in \cite{mao2017exploring}.
Besides, we observe that \framework performs much better than heuristics, which demonstrates the importance of searching for the best pattern sets via AutoML and our shrunk search space makes sense. 
Furthermore, we have the similar conclusion in Figure \ref{fig:exp1}(c).

\begin{figure}[t]
\setlength{\abovecaptionskip}{0.1cm}
\setlength{\belowcaptionskip}{-0.65cm} 
\flushleft
\includegraphics[width=1.0\linewidth]{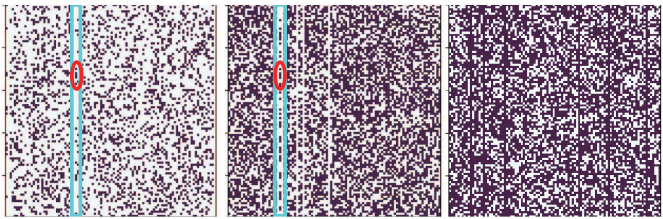}
\captionsetup{font={footnotesize}}
\caption{The illustration of patterns identified for 3 V/F levels on Transformer.}
\label{fig:visualization}
\end{figure}

\textbf{Visualization.}
{Figure \ref{fig:visualization} shows the visualization results of patterns with diverse sparsity on the self-attention layer of the first Transformer encoder. For other resultant layers, they have similar results. In these figures, the purple pixels represent 1 in the pattern, and others will be pruned from the weight. It is clear that the identified pattern sets by RL have different sparsity (e.g., 75\%, 50\%, 37\%), corresponding to different V/F levels. 
When V/F level is changed, one pattern set is swapped out to off-chip memory and another is swapped in from off-chip memory. We have an interesting observation, for Figure \ref{fig:visualization}(a) and (b), the blue box shows the similar column characteristic.
In addition, the circle part has exactly the same shape, which reflects that the proposed search space generation approach based on BP can find the important positions for higher accuracy.
}

\subsection{\framework: Ablation Study}

To further evaluate \framework, we carry out two sets of ablation studies on Transformer and DistilBERT.

\begin{figure}[t]
\setlength{\belowcaptionskip}{-0.6cm} 
\flushleft
\includegraphics[width=1.0\linewidth]{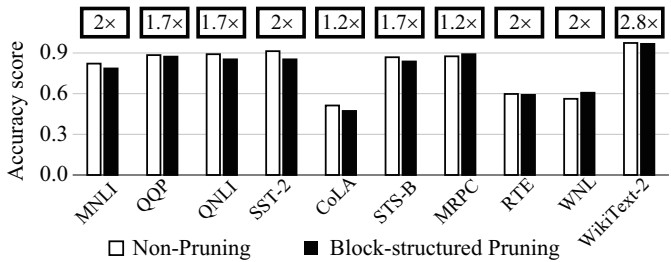}
\captionsetup{font={footnotesize}}
\caption{Evaluation of the block-structured pruning.}
\label{fig:exp_4_3_1}
\end{figure}

\textbf{Block-structured Pruning (\textit{BP}).}
Figure \ref{fig:exp_4_3_1} reports the comparison results in the scores of DistilBERT \cite{sanh2019distilbert} on 9 GLUE benchmark tasks and the accuracy of Transformer on WikiText-2.
In this figure the y-axis is the scores/accuracy, and the white bars stand for the score of original model and the black ones for ``\textit{BP}''.
The numbers in the rectangles are the pruning rate achieved by \textit{BP}.
From the figure, it is clear that \textit{BP} can achieve up to $2\times$ compression ratio with only 1.74\% accuracy loss on average. From the results, BP as the first-step pruning guarantees the accuracy and pruning rate.

Table \ref{tab:ablation study} reports the comparison of \framework against the random approach.
In Table \ref{tab:ablation study}, columns ``Non-Opt'', ``rBP only'', and ``BP only'' report the results of original model, the random approach and block-structured pruning, respectively, among which \textit{rBP} conducts pruning by randomly select rows/columns in blocks.
It is clear to see that the proposed \textit{BP} can reduce the model size in a better way, where \textit{BP} and \textit{rBP} have almost the same number of runs, while the accuracy loss of \textit{BP} is only 0.64\%, but it is 2.03\% for \textit{rBP} for WikiText-2 dataset. The same conclusion is conducted in RTE and STS-B tasks.

\textbf{AutoML based Pattern Pruning (\textit{PP}).} From Table \ref{tab:ablation study}, ``rBP+rPP'' and ``rBP+PP'' use the random-based \textit{rPP} and the proposed \textit{PP}.
Similar to \textit{rBP}, \textit{rPP} randomly determine the pattern sets and the detailed patterns in each set.
From the results of WikiText-2, it is clear that \textit{rBP+rPP} suffers 11.07\% accuracy loss, while only 4.88\% for \textit{rBP+PP}. It indicates that our \textit{PP} can retain feature information better because of the search space generation method.
Furthermore, \framework using \textit{BP+PP} can obtain results with merely 0.95\% accuracy loss, meanwhile, it achieves 4.96$\times$ improvement on the number of runs, prolonging the battery lifetime. For RTE and STS-B tasks on DistilBERT, the similar results are showed in Table \ref{tab:ablation study}.

All the above results demonstrate that \framework can guarantee accuracy, pruning rate and hardware efficiency  by using hybrid BP and PP creatively. Moreover, coupling hardware reconfiguration with software reconfiguration can prolong battery lifetime best and satisfy real-time performance.  


\section{Conclusion}
This work presented an pruning-based AutoML framework for reconfigurable Transformers, aiming at accommodating the large-size models to resource-constrained mobile platforms with real-time reconfiguration to prolong usage of battery-powered devices.
\framework utilizes a hybrid block-structured pruning and pattern pruning to deploy Transformers. And it combines hardware reconfiguration (i.e., DVFS) and software reconfiguration (i.e., multiple pattern sets) to support run-time reconfigurability.
Results show that \framework can prolong battery lifetime for 4.96$\times$ and maintain the model accuracy/score with only 0.95\% loss on Transformer and 1.5\% on DistilBERT.

\scriptsize
\bibliographystyle{ieeetr}
\bibliography{ref}

\end{document}